\documentclass[letterpaper]{article} 
\usepackage{aaai2026}  
\usepackage{times}  
\usepackage{helvet}  
\usepackage{courier}  
\usepackage[hyphens]{url}  
\usepackage{graphicx} 
\usepackage{natbib}  
\usepackage{caption} 
\usepackage{algorithm}
\usepackage{algorithmic}
\usepackage{amsmath}
\usepackage{booktabs}
\usepackage{multirow}

\usepackage{newfloat}
\usepackage{listings}
\DeclareCaptionStyle{ruled}{labelfont=normalfont,labelsep=colon,strut=off} 
\floatstyle{ruled}
\newfloat{listing}{tb}{lst}{}
\floatname{listing}{Listing}
 \nocopyright 

\title{Uncertainty Under the Curve: A Sequence-Level Entropy Area Metric for Reasoning LLM}
\author {
    Yongfu Zhu,
    Lin Sun,
    Guangxiang Zhao,
    Weihong Lin,
    Xiangzheng Zhang
}
\affiliations {
    Qiyuan Tech
}

\usepackage{bibentry}

\begin{document}

\maketitle

\begin{abstract}
In this work, we introduce Entropy Area Score (EAS), a simple yet effective metric to quantify uncertainty in the answer generation process of reasoning large language models (LLMs). EAS requires neither external models nor repeated sampling, it integrates token-level predictive entropy from the model itself to capture the evolution of uncertainty during generation. Empirical results show that EAS is strongly correlated with answer entropy across models and datasets. In training data selection, EAS identifies high-potential samples and consistently outperforms Pass Rate filtering under equal sample budgets, improving student model accuracy on math benchmarks. EAS is both efficient and interpretable, offering a practical tool for uncertainty modeling and data quality assessment in LLM training.
\end{abstract}


\section{1~ ~Introduction}

Reasoning LLMs have shown strong performance in complex domains such as mathematics and science. However, their outputs remain sensitive to minor changes in evaluation conditions (e.g., random seeds, temperature, prompt format), leading to significant fluctuations in reported scores and undermining reproducibility \cite{hochlehnert2025soberlookprogresslanguage, sun2025evaluationneedstrategicoverclaiming}.

These fluctuations often stem from the model’s uncertainty when solving ambiguous or borderline problems. Stable results occur only when a model consistently answers correctly or incorrectly; in contrast, variance across runs reveals internal indecision. This highlights the need for reliable, fine-grained methods to quantify model uncertainty during the reasoning process—both to improve evaluation stability and to support downstream tasks such as training data selection.

To address this, we propose Entropy Area Score (EAS), a simple and efficient metric that tracks the evolution of uncertainty throughout generation and provides actionable insights into model behavior.

\subsection{1.1~ ~Related Work}
Existing approaches for modeling and quantifying LLM uncertainty can be broadly classified into two categories: 

\subsubsection{I. Explicit uncertainty estimation through model-internal or auxiliary mechanisms.}These methods typically involve training the model itself or an external scoring model to estimate output uncertainty or confidence levels. For example, \citet{lin2022teachingmodelsexpressuncertainty} fine-tunes GPT-3 to produce not only answers but also natural-language confidence estimates (e.g., 90\% confidence). Similarly, \citet{kadavath2022languagemodelsmostlyknow} adds an auxiliary confidence head to the model and trains it via confidence-based fine-tuning. Other approaches, such as \citet{tian-etal-2023-just}, elicit confidence estimates either directly or through a two-stage prompting process.

Beyond model-internal confidence heads, some works incorporate external probing modules. For instance, \citet{heo2025llmsestimateuncertaintyinstructionfollowing} trains a linear classifier to map internal LLM representations to task success, treating the classifier’s probability as an uncertainty score. \citet{liu2024uncertaintyestimationquantificationllms} uses supervised learning on labeled data, leveraging both hidden activations and output entropy to predict the model’s confidence.

While effective in some settings, these methods rely heavily on either model quality or external supervision. As noted by \citet{NEURIPS2024_9c20f16b}, lower-performing models tend to produce overconfident yet incorrect predictions, limiting the reliability of self-reported confidence. Moreover, training task-specific uncertainty models is costly and lacks generalizability across domains and model scales.

\subsubsection{II. Implicit uncertainty estimation via statistical signals from model outputs.}This category avoids training additional components and instead uses statistical properties—e.g., token-level log probabilities or entropy—to assess uncertainty. A classical metric is perplexity  \cite{10.1121/1.2016299}, proposed early on as a measure of linguistic difficulty and prediction uncertainty. In machine translation, \citet{10.1162/tacl_a_00330} leverages output probabilities and attention weights to calibrate confidence. \citet{10.1162/tacl_a_00324} focuses on multiple-choice QA and analyzes signals such as the margin between top predicted options to evaluate answer confidence.

\citet{Farquhar2024} samples multiple outputs and uses entropy over semantic equivalence to quantify uncertainty. These approaches have inspired our work by highlighting how repeated outputs or token-level statistics can reflect model behavior.

In reasoning-intensive benchmarks like mathematics or science, answers are typically unique and unambiguous, which makes these domains ideal for studying model uncertainty. Moreover, we observe that powerful reasoning LLMs like DeepSeek-R1-0528 frequently generate tokens such as “Wait”, “But”, or “Alternatively” during math problem-solving—indicative of a dynamic trial-and-error process. For instance, in a random sample of 100K math examples from AM-DeepSeek-R1-0528-Distilled \cite{AM-DeepSeek-R1-0528-Distilled}, the combined frequency of these three tokens is 0.98\%, i.e., roughly once every 100 tokens.

Such patterns reveal that reasoning is not a static process but one involving continual hypothesis revision. Yet, many uncertainty metrics only consider output-final statistics (e.g., entropy of the last token), failing to capture the trajectory of uncertainty as the model thinks and generates.

\subsection{1.2~ ~Our Contribution}
To address these limitations, we propose a new metric: Entropy Area Score (EAS), which explicitly models the evolution of token-level uncertainty across the generation path. Our goal is to measure how confident the model is in its own answer—not how correct the answer is which aligns with the concept of distributional uncertainty (i.e., “not knowing what it doesn't know”) discussed in \cite{NEURIPS2018_3ea2db50}.

Although it does not directly measure answer correctness, such uncertainty quantification remains valuable—for instance, in training data selection. A popular strategy, seen in works like \citet{lyu2025exploringlimitoutcomereward} uses Pass Rate filtering: only retaining examples with intermediate accuracy (neither all-correct nor all-wrong) during multi-sample reasoning. This approach has proven effective across multiple reasoning LLMs training \cite{kimiteam2025kimik2openagentic, deepseekai2025deepseekr1incentivizingreasoningcapability, wen2025lightr1curriculumsftdpo}.

However, it suffers from logical limitations and high computational cost due to repeated sampling. In contrast, our proposed EAS only requires a single forward pass and consistently outperforms baseline strategies in both accuracy and efficiency across architectures and model sizes.

In summary, our key contributions are as follows:
\begin{enumerate}
    \item \textbf{EAS: A novel metric for modeling uncertainty in language model outputs.} EAS is simple to compute, requiring no auxiliary models or fine-tuning. It directly leverages the model’s native token-level predictions, enabling generalization across tasks and models with minimal cost. Moreover, it provides a dynamic trajectory of uncertainty, offering fine-grained interpretability.
    \item \textbf{Demonstrated correlation with sampling-based uncertainty.} We show that EAS strongly correlates with answer entropy derived from repeated generation across multiple models and tasks, validating it as a reliable proxy for output uncertainty.
    \item \textbf{Effective application in data selection for training.} By identifying samples where the model exhibits high uncertainty during generation, EAS helps select data with high learning potential. Compared with random, length-based, or Pass Rate-based selection, EAS consistently improves student model performance under the same sample size, showing its practical utility in large-scale training pipelines.
\end{enumerate}

\section{2~ ~Entropy Area Score (EAS)}

\begin{figure}[H]
    \centering
    \includegraphics[width=0.96\linewidth]{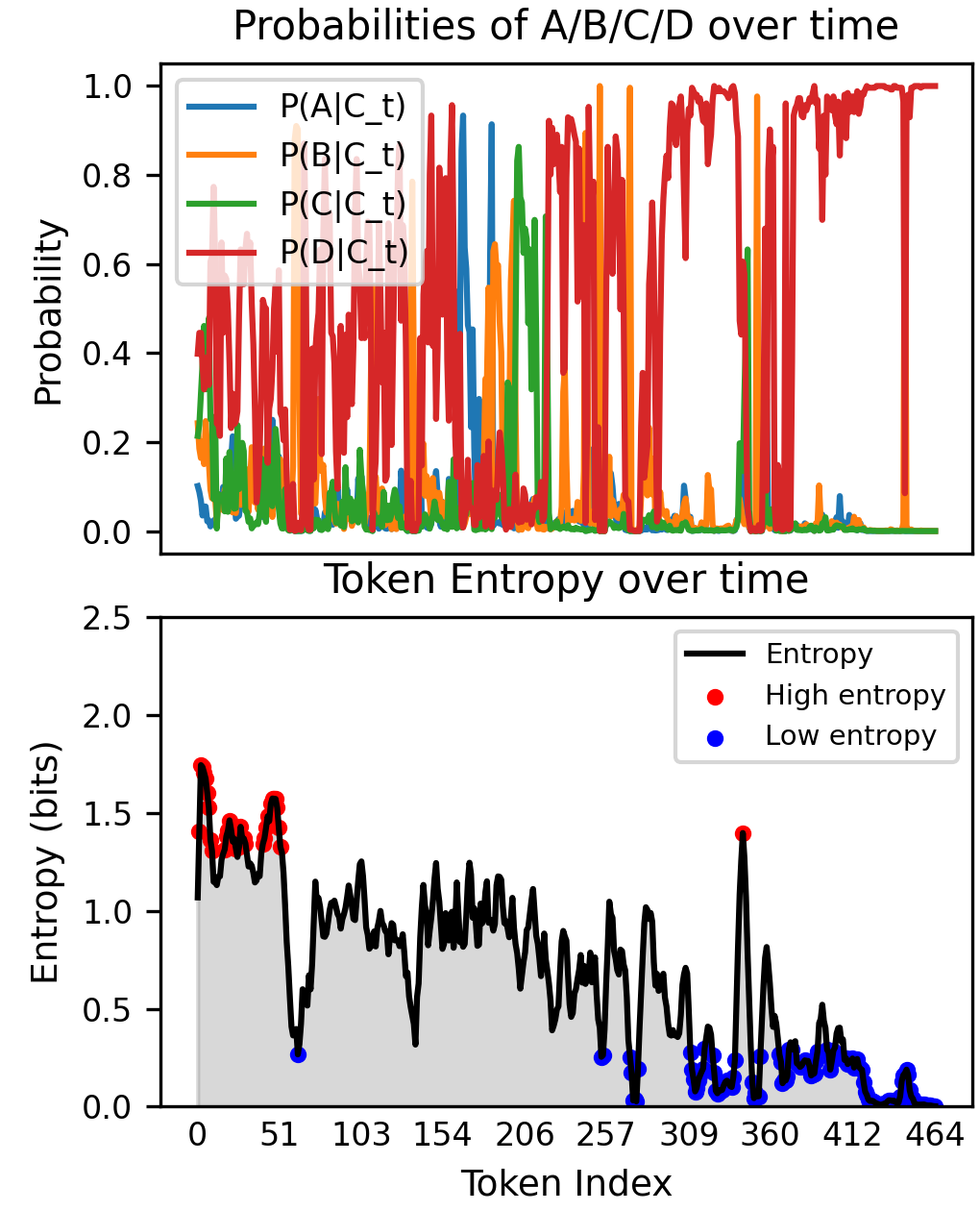}
    \caption{\textbf{Token-Level Entropy Trajectory and EAS Computation.} (1) The shaded area under the entropy curve represents the EAS score, reflecting cumulative uncertainty. (2) In the early stages, the model shows no clear preference, leading to higher entropy. As generation progresses and the model converges on a likely answer, entropy decreases.}
    \label{fig: figure1}
\end{figure}

To quantify uncertainty in the answer generation process of reasoning models, especially in domains like mathematics and science, we formally define the EAS as follows:

\begin{itemize}
    \item \textbf{Step 1: Context Construction.} Let the model’s generated token sequence be $S = (x_1, x_2, \dots, x_T, \dots)$, where $x_T$ is the last token of the final answer (note that it may not be the last token of the full sequence). At each position $t \in \{1, 2, \dots, T-1\}$, we construct a context $\tilde{C}_t$ for predicting the next token as:
    \[
    \tilde{C}_t = (x_1, x_2, \dots, x_t,\ 
    \mathtt{\text{``\textbackslash boxed\{''}},\ 
    \mathtt{prefix}_{\text{ans}}^{(L-1)})
    \]
    \item \textbf{Step 2: Entropy Computation.} At each position $t$, the model produces a predictive distribution over the vocabulary $\mathcal{V}$: $P_t(v) = P(v \mid \tilde{C}_t), \quad v \in \mathcal{V}$.
    The corresponding token-level entropy is then defined as:
    \[
    \mathcal{H}t = - \sum_{v \in \mathcal{V}} P_t(v) \log_2 P_t(v)
    \]
    \item \textbf{Step 3: Area Integral of Entropy (EAS).} We compute the sum of entropy values from position $1$ to $T-1$ to capture the total uncertainty across the generation trajectory:
    \[
    \text{EAS}(S) = \sum_{t=1}^{T-1} \mathcal{H}_t
    \]
\end{itemize}

\section{3~ ~Comparison with Other Metrics}

To assess the effectiveness of EAS, we compare it against other commonly used and lightweight uncertainty metrics.

\begin{figure*}[t]
    \centering
    \includegraphics[width=0.96\linewidth]{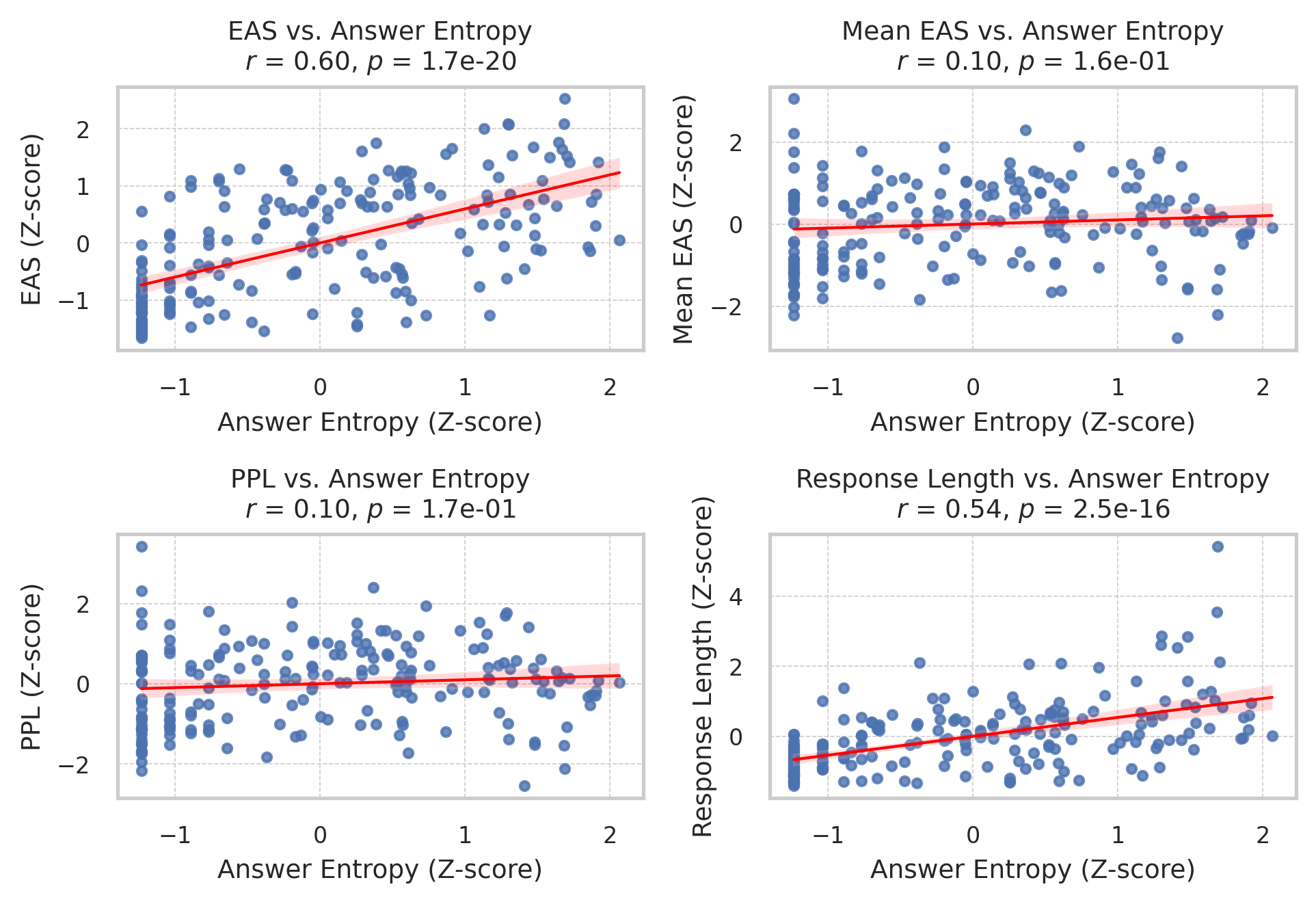}
    \caption{\textbf{Correlation Between Uncertainty Metrics and Answer Entropy.} Each point represents a GPQA-Diamond question. Metrics are Z-score normalized. The red line indicates linear regression; Pearson’s r and p-value are shown per subplot. EAS shows the strongest correlation, indicating it best captures model uncertainty.}
    \label{fig: figure2}
\end{figure*}

\subsection{3.1~ ~Experimental Setup}

We run 64 inferences with DeepSeek-R1-Distill-Qwen-14B on the GPQA-Diamond dataset (198 science questions) to compute sampling-based uncertainty. A standard metric is correctness entropy, which treats each output as correct or incorrect and measures uncertainty as:
\[
\mathcal{H}_{\text{correctness}} = -p \log_2 p - (1 - p) \log_2 (1 - p)
\]
where $p = \frac{n_{\text{c}}}{N}$ is the proportion of correct answers among the $N=64$ samples.

However, this binary view lacks granularity—it cannot distinguish between consistently wrong predictions and varied incorrect answers. Since we aim to capture the model’s internal uncertainty, we instead use answer entropy, which reflects the diversity of generated answers:

Let $\mathcal{A} = \{ a_1, a_2, \dots, a_K \}$ denote the set of unique answers generated by the model across the 64 runs, and $n_k$ be the count of answer $a_k$. The answer entropy is then computed as:
\[
\mathcal{H}_{\text{answer}} = - \sum_{k=1}^{K} p_k \log_2 p_k \text{,} \quad p_k = \frac{n_k}{N}
\]
This reference metric reflects how diverse or unstable the model’s outputs are under repeated inference. Low answer entropy indicates consistent outputs; high entropy reveals indecision or instability.

To evaluate different uncertainty metrics, we compute each metric per question and report its Pearson correlation with answer entropy across all 198 samples. This correlation quantifies how well each metric approximates sampling-based uncertainty. We compare the following metrics:
\begin{itemize}
    \item \textbf{EAS: }Computed via a single forward pass, EAS integrates token-level entropy across the generation path. At each step, we construct a context suffix including \textit{``\textbackslash boxed\{''} and the first $L-1$ tokens of the ground-truth answer, and use the vLLM \cite{kwon2023efficient} API to extract top-K token probabilities. To account for truncation, we estimate the maximum entropy error where V = $151{,}665$ and $K = 20$ as:
    \[
    \Delta \mathcal{H}{\max} \leq \varepsilon \cdot \log_2 \left( \frac{V - K}{\varepsilon} \right), \quad \varepsilon = 1 - \sum_{i=1}^{K} p_i
    \]

    \begin{table}[htbp]
        \centering
        \begin{tabular}{ccc}
            \toprule
            \multirow{2}{*}{Mertic} & Total Probability & Theoretical  \\
            & Mass of Top-K & Error Bound  \\
            \midrule
            All Tokens Average & 99.87\% & 0.031 \\
            90th Percentile Average & 99.85\% & 0.040 \\
            95th Percentile Average & 99.52\% & 0.12 \\
            99th Percentile Average & 97.54\% & 0.56 \\
            \bottomrule
        \end{tabular}
        \caption{\textbf{Upper Bound of Entropy Truncation Error in EAS.} Estimated maximum error from truncating the token distribution to top-K (K=20), based on the principle of maximum entropy. Results show minimal distortion, validating the approximation used in EAS.}
        \label{tab: table1}
    \end{table}

    As shown in Table~\ref{tab: table1}, top-20 tokens capture over 99.87\% of probability mass on average, limiting entropy error to below 0.031 bits. Given average entropy is approximately 0.66, this approximation introduces negligible distortion less than 4.70\%. These findings confirm that using a truncated probability distribution for EAS offers a sound and robust approximation while greatly improving computational efficiency.

    \item \textbf{Mean Entropy Area Score (Mean EAS): }This is simply the average entropy across token positions:
    \[
    \text{Mean EAS}(S) = \frac{1}{T-1} \sum_{t=1}^{T-1} \left( - \sum_{v \in \mathcal{O}} P_t(v) \log_2 P_t(v) \right)
    \]

    \item \textbf{Perplexity (PPL): }This metric quantifies the model’s perplexity over its own generated token sequence, reflecting the overall uncertainty in its predictions:
    \[
    \text{PPL}(S) = \exp \left( -\frac{1}{|S|} \sum_{t=1}^{|S|} \log P(x_t \mid x_{<t}) \right)
    \]
    \item \textbf{Response Length: }The total number of tokens $|S|$ in the model’s response sequence.
    
\end{itemize}

\subsection{3.2~ ~Experimental Results and Analysis}

\begin{figure*}[t]
    \centering
    \includegraphics[width=0.96\linewidth]{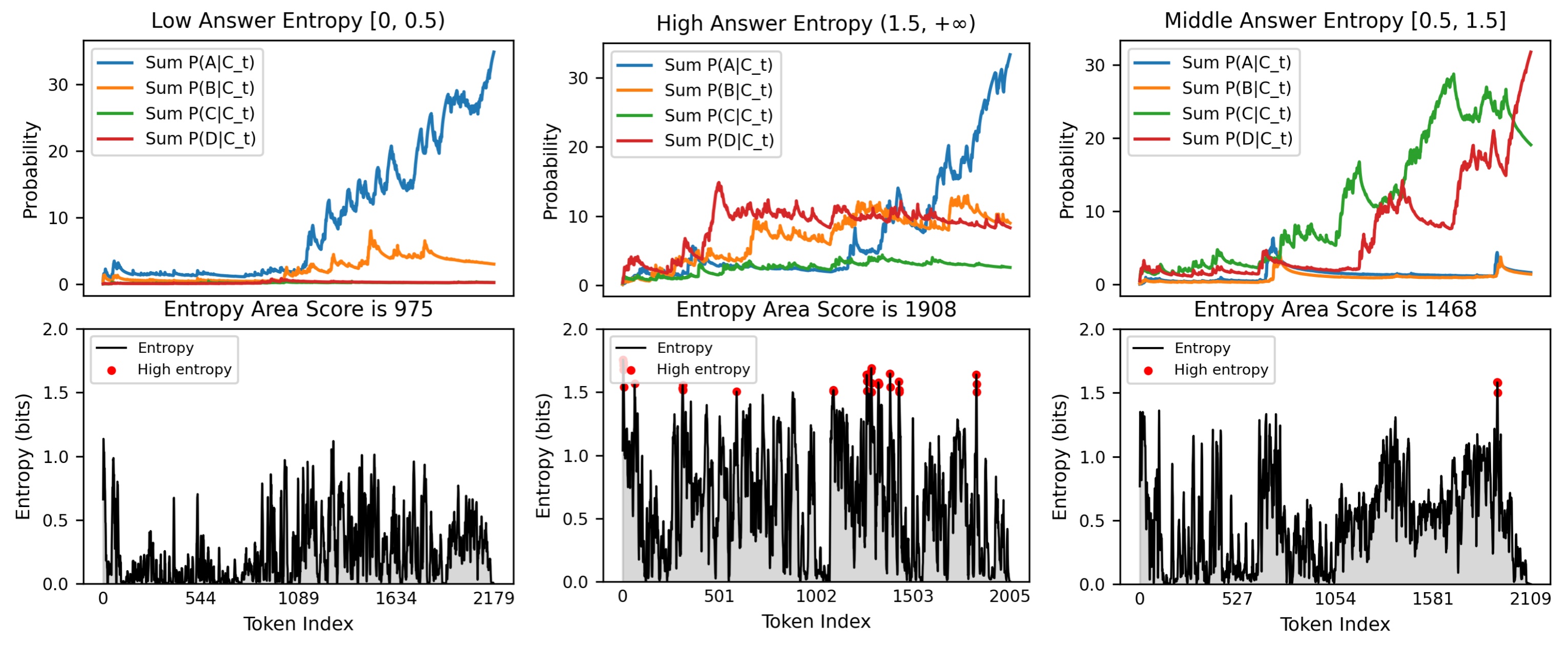}
    \caption{\textbf{Model Behavior Across Different Levels of Answer Entropy.} Questions are grouped by answer entropy: low $[0, 0.5)$, medium $[0.5, 1.5]$, and high $(1.5, +\infty)$, with sample counts 60:97:51. (1) Low-entropy examples show early and stable preference for the correct option. (2) High-entropy cases exhibit frequent option switches and fluctuating uncertainty. (3) Medium-entropy examples display partial stability followed by revision. Token-level entropy curves confirm that earlier and wider uncertainty leads to higher EAS scores. These trends explain the strong alignment between EAS and answer entropy.}
    \label{fig: figure3}
\end{figure*}

We visualize the correlation between answer entropy and four uncertainty metrics—EAS, Mean EAS, PPL, and Response Length—in Figure~\ref{fig: figure2}. Among them, EAS shows the strongest linear correlation, suggesting it best reflects model uncertainty under repeated inference. This is because answer entropy captures the consistency of model preferences across runs, while EAS integrates uncertainty across the entire generation path.

To enhance prediction salience and reduce noise from low-probability tokens, we append the special suffix \textit{``\textbackslash boxed\{''} to guide answer generation. To further interpret EAS’s behavior, we introduce a visualization based on decayed cumulative option probabilities, which reveals how the model’s preference over A/B/C/D evolves during generation. This method highlights the model’s internal decision dynamics in multiple-choice settings.

We begin by formally defining the decayed cumulative probability formulation that enhances the interpretability of model preference trajectories.

\begin{enumerate}
    \item At each generation step $t$, let the model’s predicted probability distribution over the multiple-choice options be 
    $P_t^{(\mathcal{O})} = \{ P_t(A), P_t(B), P_t(C), P_t(D) \}$. This gives us a snapshot of the model’s preference over the answer options at time t, based on the context prefix $\tilde{C}_t$ defined earlier. However, directly plotting $P_t^{(\mathcal{O})}$ across time results in highly oscillatory curves due to sensitivity to the immediate context. These fluctuations obscure the true evolution of preference, as temporary alignment between the context and a specific option token can artificially spike its predicted probability.
    
    \item To smooth the noise and highlight overall trends, we compute the cumulative sum of predicted probabilities for each option up to step $t$:
    \[
    \widehat{P}_t^{(\mathcal{O})}(v) = \sum_{k=1}^{t} P_k(v), \quad \forall v \in \{A, B, C, D\}
    \]
    This formulation suppresses short-term volatility and emphasizes which option the model consistently leans toward. However, it treats all positions equally and may mask the influence of recent evidence—especially when the model shifts its preference mid-generation. Furthermore, since each token’s value becomes a first-order difference of the cumulative curve, it is harder to observe relative probability changes among competing options.

    \item To capture both the smoothness and time sensitivity of model preferences, we define a decayed cumulative probability with distance-based weighting: 
    \[
    \widetilde{P}t^{(\mathcal{O})}(v) = \sum_{k=1}^{t} \frac{1}{(t - k + 1)^\alpha} \cdot P_k(v)
    \]
    
    Here, $\alpha > 0$ is a decay coefficient that controls how much weight is assigned to earlier positions. When $\alpha = 0$, this reduces to the unweighted cumulative probability (all positions equally weighted); when $\alpha \to +\infty$, it reduces to the raw one-step prediction $P_t^{(\mathcal{O})}$, focusing entirely on the current token.
    By setting $\alpha = 0.5$, we found an effective trade-off between smoothing and local responsiveness. This value ensures that the final answer’s curve is typically dominant at the last token (i.e., the model’s prediction is visually validated), while preserving readable trends over time.
    
\end{enumerate}

We then discretize the questions based on their answer entropy using a decay coefficient of 0.5, and examine intra-bucket similarities and inter-bucket differences. As illustrated in the upper portion of Figure~\ref{fig: figure3}, we observe several consistent behavioral patterns across entropy levels:

\begin{itemize}
    \item \textbf{Observation 1: The model exhibits a strong and stable preference for the final answer option in low-entropy samples.} In the $[0, 0.5)$ group, the model quickly commits to the final answer, with its probability curve rising early and remaining dominant. Competing options show only brief, minor fluctuations. This reflects strong internal confidence and results in consistent outputs across runs. However, such confidence may be misplaced: 4 of the 60 questions in this group scored 0, indicating overconfident hallucination.

    \item \textbf{Observation 2: No option maintains a consistent lead in high-entropy examples.} In the $(1.5, +\infty)$ group, the model’s preference shifts frequently, with option curves intersecting throughout the sequence. This indicates indecision and sensitivity to small perturbations. The resulting outputs vary widely across runs, and only 4.4\% of these questions exceeded the average Pass@1 score of 57.64\%.

    \item \textbf{Observation 3: Medium-entropy samples displays mixed dynamics.} In the $[0.5, 1.5]$ group, one option often shows an early lead, but is later overtaken by another option. The model may initially favor a plausible answer, then revise its judgment after generating additional context or reaching a key reasoning step. Compared to low-entropy samples, the turning point where the dominant curve emerges typically occurs later. Compared to high-entropy samples, the number of curve intersections is fewer, and full tie scenarios (where no curve dominates) are rare. These samples reflect partial stability with local revision, where the model starts confidently but later shifts its hypothesis—a behavior that aligns with mid-range answer entropy.

\end{itemize}

From the above case studies, we can conclude that a lower answer entropy is typically observed when the leading option curve begins rising earlier in the generation process, exhibits denser upward segments, and has fewer intersections with other option curves. 

As shown in the lower part of Figure~\ref{fig: figure3}, these characteristics are directly reflected in the token-level entropy curves: low-entropy examples tend to have longer segments where entropy remains low, while high-entropy examples display sharp increases earlier in the sequence and across a broader range. Consequently, the total accumulated uncertainty—quantified by the EAS—is smaller for low-entropy samples and larger for high-entropy samples, reinforcing the observed positive correlation between EAS and answer entropy.

To understand why EAS outperforms other uncertainty metrics in capturing this behavior, we compare it against the remaining baselines:

\begin{itemize}
    \item \textbf{Comparison to Mean EAS:} Mean EAS, which simply averages the token-level entropy across the sequence, fails to distinguish medium-entropy cases where the model initially shows strong preference for one option but later reverses course. In such cases, the entropy at each position may remain low because the model maintains confidence at each step—even though its final answer is unstable across runs. This leads to Mean EAS underestimating uncertainty, and in some cases assigning lower scores to medium-entropy examples than to truly stable (low-entropy) ones. In contrast, EAS, as an integral over the entire entropy curve, captures not just the momentary uncertainty but also the extent and duration of uncertainty over time, effectively distinguishing these cases.

    \item \textbf{Comparison to PPL: } PPL measures the model’s uncertainty in language generation at the token level. A low PPL indicates high fluency or grammatical consistency, but not necessarily high confidence in the semantic correctness of the answer. In other words, a model may generate a smooth, well-formed response while still being uncertain about the answer. Hence, PPL and answer entropy are not causally linked, and the correlation between PPL and uncertainty is weak in our experiments.

    \item \textbf{Comparison to Response Length: }Response length is sometimes used as a proxy for uncertainty, assuming longer outputs indicate more reasoning. However, this correlation is inconsistent and task-dependent. In GPQA-Diamond, one organic chemistry question led to outputs exceeding 6,300 tokens—longer than 6\% of samples—yet achieved 100\% Pass@1 with zero answer entropy. This length was due to long chemical formulas (e.g., $C_6H_{12}O_2$ being tokenized into 7 subwords). In such cases, response length reflects input structure rather than uncertainty. Overall, its correlation with answer entropy is weaker and noisier than that of EAS.
\end{itemize}

EAS consistently shows the strongest correlation with sampling-based answer entropy, as it captures both local hesitation and global uncertainty across the generation process. This makes it a reliable and interpretable proxy for model uncertainty in complex reasoning tasks.

\subsection{3.3~ ~Generalization Across Models and Tasks}

In addition to the experiments above, we further evaluate the generality and robustness of the EAS metric under varying model scales, architectures, and task types. It is worth noting that all selected models and tasks fall within the domain of reasoning, which aligns with the objective of this study. 

The evaluated models include various sizes and architectures:

\begin{itemize}
    \item \textbf{Model families:} Qwen2.5 and LLaMA
    \item \textbf{Parameter scales:} Ranging from 8B to 14B
\end{itemize}

The selected tasks are:
\begin{itemize}
    \item \textbf{AIME24 and AIME25:} Two mathematics competitions benchmarks, which we merge and report as a combined set
    \item \textbf{GPQA-Diamond: } A science-based QA dataset
\end{itemize}

\begin{table}[htbp]
    \centering
    \begin{tabular}{ccc}
        \toprule
        \multirow{1}{*}{Model} & Average AIME & GPQA-Diamond \\
        \midrule
        DS-R1-Qwen-14B & 0.8237 & 0.5968 \\
        DS-R1-LLaMA-8B & 0.6820 & 0.5434 \\
        \bottomrule
    \end{tabular}
    \caption{\textbf{Correlation Between EAS and Answer Entropy Across Models and Tasks.} Pearson correlation coefficients for EAS vs. answer entropy on AIME and GPQA-Diamond benchmarks using different DeepSeek-R1-Distill models. All correlations are statistically significant ($p < 5e-5$). Results show that EAS maintains strong correlation across model scales and architectures.}
    \label{tab: table2}
\end{table}

The results are summarized in Table~\ref{tab: table2}. These results confirm that EAS maintains strong and significant correlation with answer entropy across different architectures, parameter sizes, and reasoning tasks. This suggests that EAS is a stable and generalizable uncertainty metric for evaluating large language models on complex reasoning problems.

\section{4~ ~Training Data Selection Based on EAS}

Having established EAS as a reliable uncertainty metric, we explore its utility in training data selection—a critical challenge in LLM training, where noisy or low-quality samples can waste resources and hinder performance. By modeling uncertainty across the generation trajectory, EAS helps distinguish between easy, hard, and ambiguous examples, making it suitable for filtering and curriculum design. We validate this through an ablation study under controlled settings.

\subsection{4.1~ ~Experimental Setup}

\begin{table*}[ht]
    \centering
    \begin{tabular}{cccccc}
        \toprule
        \multirow{1}{*}{Model} & Checkpoint-Rank & Random Sampling & Length-Based & Pass Rate-Based & EAS-Based  \\
        \midrule
        \multirow{3}{*}{DS-R1-LLaMA-8B} & 1 & 56.8 & +1.4 & -2.5 & \textbf{+2.1} \\
        & 2 & 56.6 & +1.5 & -3.3 & \textbf{+2.3} \\
        & 3 & 56.3 & \textbf{+1.7} & -3.3 & +1.5 \\
        \midrule
        \multirow{3}{*}{DS-R1-Qwen-14B} & 1 & 77.0 & +0.3 & +0.7 & \textbf{+1.2} \\
        & 2 & 76.7 & +0.4 & +0.8 & \textbf{+1.2} \\
        & 3 & 76.4 & +0.7 & +0 & \textbf{+1.3} \\ 
        \bottomrule
    \end{tabular}
    \caption{\textbf{Performance Comparison of Different Data Selection Strategies.} Average Pass@1 scores on AIME24 and AIME25 using four data selection strategies. Each model reports results from its top 3 validation checkpoints. EAS-based selection consistently achieves the best performance, highlighting its effectiveness in identifying high-potential training samples via single-pass uncertainty estimation.}
    \label{tab: table3}
\end{table*}

We evaluate the effectiveness of EAS-based data selection in a supervised fine-tuning (SFT) setup, controlling data selection as the only variable. Experiments are conducted on two base models: DeepSeek-R1-Distill-Qwen-14B (DS-R1-Qwen-14B) and DeepSeek-R1-Distill-LLaMA-8B (DS-R1-LLaMA-8B), using the math subset of AM-DeepSeek-R1-0528-Distilled, which has already undergone correctness verification, de-duplication, and contamination filtering. We further apply vector-based contamination filtering and remove samples exceeding 20,480 tokens.

In the data selection strategies, we compared:

\begin{itemize}
    \item \textbf{Random Sampling:} Uniformly sample 5,000 examples as a baseline and ensures equal sample size across strategies to eliminate size-based bias.
    \item \textbf{Length-Based Selection:} Sort samples in descending order by total token length, motivated by the observation (e.g., in OpenAI-o1, DeepSeek-R1) that longer samples often include more complex context, semantic structures, or reasoning chains, potentially contributing more to model training.
    \item \textbf{Pass Rate-Based Selection:} For each training sample, we conduct 4 rounds of repeated inference and compute Pass Rate as the ratio of correct outputs. Then we filter out samples with Pass Rate = 1 (too easy) or 0 (too hard). Among the remaining samples, retain those with lowest Pass Rates, which are hypothesized to be challenging but learnable.
    \item \textbf{EAS-Based Selection:} For each training sample, run one forward pass using the base model and compute its EAS. Then we retain the top 5,000 samples with the highest EAS scores and these are considered the samples where the model showed the most internal uncertainty during generation, indicating potential learning value.
\end{itemize}

All selected samples are trained using identical hyperparameters under the ms-swift framework \cite{zhao2024swiftascalablelightweightinfrastructure}. For evaluation, we use the evalscope \cite{evalscope_2024} framework to measure performance on the AIME24 and AIME25 datasets, each with 32 rounds of repeated inference. The metric reported is Average Pass@1, using the same inference parameters as the original base models.

\subsection{4.2~ ~Experimental Results and Analysis}

The results in Table~\ref{tab: table3} demonstrate that EAS-based data selection consistently achieves the best performance across all checkpoints and outperforms both Pass Rate and Length-based strategies.

We attribute EAS’s advantage over Pass Rate to the following key factors:

\subsubsection{I. EAS provides a finer-grained, single-pass uncertainty estimate.} Pass Rate depends on multiple rounds of repeated inference and is constrained by discrete values (e.g., 0, 0.25, 0.5, 0.75, 1 when using 4 samples). This coarse granularity makes it hard to distinguish between subtly different samples. For instance, a sample where the model wavers between correct and incorrect options may still be categorized the same as a highly confident one.

Moreover, the common practice of filtering out Pass Rate = 1 and 0 samples and selecting those with low but non-zero Pass Rates introduces logical inconsistencies. In our earlier evaluation, among the 198 GPQA-Diamond questions, 44 had all 4 initial responses incorrect—yet 9 of them produced at least 1 correct answer in the next 4 runs. These cases, under Pass Rate = 4 setting, would be wrongly discarded as “unsolvable,” despite having actual learning value.

By contrast, EAS is computed from a single forward pass, is continuous-valued, and captures token-level uncertainty across the entire output trajectory. It distinguishes between consistently wrong-but-confident samples and those with meaningful internal struggle—i.e., the “high-potential” samples.

\subsubsection{II. EAS focuses on model–data interaction, not just model correctness.} 
Pass Rate only measures whether the model gets the right answer during repeated inference. It reflects outcome, but not the internal process or pedagogical value of the sample. 

EAS, in contrast, reflects how much uncertainty the model experiences while generating the answer. It answers: “Did this example make the model think?” rather than “Did the model get it right?”

This allows EAS to identify training examples that stimulate discriminative reasoning, contain ambiguity, or trigger hypothesis revision—qualities that are valuable for learning. In this sense, EAS can be interpreted as a model-data compatibility score, which helps prioritize samples that are neither too easy nor too hard, but rich in learning signals.

\section{5~ ~Discussion}
In this paper, we proposed Entropy Area Score (EAS) as a metric to quantify the uncertainty exhibited by large reasoning models during answer generation. Compared to existing approaches, EAS offers both computational efficiency and interpretability. We demonstrated its ability to capture shifts in model preference and to identify uncertain samples across complex reasoning tasks in mathematics and science.

However, EAS is inherently designed to measure the distributional uncertainty over answer tokens by integrating token-level entropy during generation. This design makes it less applicable to tasks with non-unique or structure-diverse outputs, such as IFEval or LiveCodeBench, where the answer is typically a free-form text or code segment. In these tasks, correctness is judged based on semantic equivalence or functional consistency, rather than the probability of generating a specific token or option.

As a result, the local uncertainty signals captured by EAS may fail to reflect the true global uncertainty of the model in such settings, leading to a lower correlation between EAS and empirical outcome variability.

Nonetheless, we believe EAS presents a novel, lightweight, and interpretable tool for uncertainty modeling in LLMs. It performs well not only in evaluation scenarios, but also proves useful in practical applications such as training data selection and curriculum optimization. In the future, we aim to extend EAS toward a more general uncertainty modeling framework applicable to broader generation tasks beyond discrete option prediction.

\bibliography{aaai2026}

\makeatletter
\@ifundefined{isChecklistMainFile}{
  \newif\ifreproStandalone
  \reproStandalonetrue
}{
  \newif\ifreproStandalone
  \reproStandalonefalse
}
\makeatother

\ifreproStandalone
\documentclass[letterpaper]{article}
\usepackage{aaai2026}
\setlength{\pdfpagewidth}{8.5in}
\setlength{\pdfpageheight}{11in}
\usepackage{times}
\usepackage{helvet}
\usepackage{courier}
\usepackage{xcolor}
\frenchspacing

\begin{document}
\fi
\setlength{\leftmargini}{20pt}
\makeatletter\def\@listi{\leftmargin\leftmargini \topsep .5em \parsep .5em \itemsep .5em}
\def\@listii{\leftmargin\leftmarginii \labelwidth\leftmarginii \advance\labelwidth-\labelsep \topsep .4em \parsep .4em \itemsep .4em}
\def\@listiii{\leftmargin\leftmarginiii \labelwidth\leftmarginiii \advance\labelwidth-\labelsep \topsep .4em \parsep .4em \itemsep .4em}\makeatother

\setcounter{secnumdepth}{0}
\renewcommand\thesubsection{\arabic{subsection}}
\renewcommand\labelenumi{\thesubsection.\arabic{enumi}}

\newcounter{checksubsection}
\newcounter{checkitem}[checksubsection]

\newcommand{\checksubsection}[1]{%
  \refstepcounter{checksubsection}%
  \paragraph{\arabic{checksubsection}. #1}%
  \setcounter{checkitem}{0}%
}

\newcommand{\checkitem}{%
  \refstepcounter{checkitem}%
  \item[\arabic{checksubsection}.\arabic{checkitem}.]%
}
\newcommand{\question}[2]{\normalcolor\checkitem #1 #2 \color{blue}}
\newcommand{\ifyespoints}[1]{\makebox[0pt][l]{\hspace{-15pt}\normalcolor #1}}

\ifreproStandalone
\end{document}
\fi


\end{document}